\documentclass[11pt, a4paper, copyright]{shopee}

\usepackage[most]{tcolorbox}
\usepackage{amsmath, amssymb}
\usepackage{float}
\definecolor{abstractgray}{RGB}{240,240,240}

\makeatletter
\renewenvironment{abstract}{%
  \begin{tcolorbox}[
    colback=abstractgray,
    colframe=white,
    arc=6pt,
    left=20pt, right=20pt,
    top=20pt, bottom=20pt,
    width=\dimexpr\linewidth+40pt\relax,  
    enlarge left by=-20pt,                
    enlarge right by=-20pt,               
    enlarge top by=-10pt,                
    before skip=0pt,
    after skip=20pt
  ]
    \absfont
}{%
  \end{tcolorbox}
}
\makeatother

\usepackage[authoryear, sort&compress, round]{natbib}
\uselogo{} 

\title{Compass-Thinker-7B Technical Report}

\renewcommand{\today}{2025-08-12}

\author{Shopee LLM Team}

\begin{document}
\maketitle

\begin{abstract}
Recent R1-Zero-like research further demonstrates that reasoning extension has given large language models (LLMs) unprecedented reasoning capabilities, and Reinforcement Learning is the core technology to elicit its complex reasoning. 
However, conducting RL experiments directly on hyperscale models involves high computational costs and resource demands, posing significant risks.
We propose the Compass-Thinker-7B model, which aims to explore the potential of Reinforcement Learning with less computational resources and costs, and provides insights for further research into RL recipes for larger models.
Compass-Thinker-7B is trained from an open source model through a specially designed Reinforcement Learning Pipeline.
We curate a dataset of 30k verifiable mathematics problems for the Reinforcement Learning Pipeline.
By configuring data and training settings with different difficulty distributions for different stages, the potential of the model is gradually released and the training efficiency is improved.
Extensive evaluations show that Compass-Thinker-7B possesses exceptional reasoning potential, and achieves superior performance on mathematics compared to the same-sized RL model.
Especially in the challenging AIME2024 evaluation, Compass-Thinker-7B achieves 40\%  accuracy.

\end{abstract}


\begin{figure}[ht]
    \includegraphics[width=1.0\textwidth]{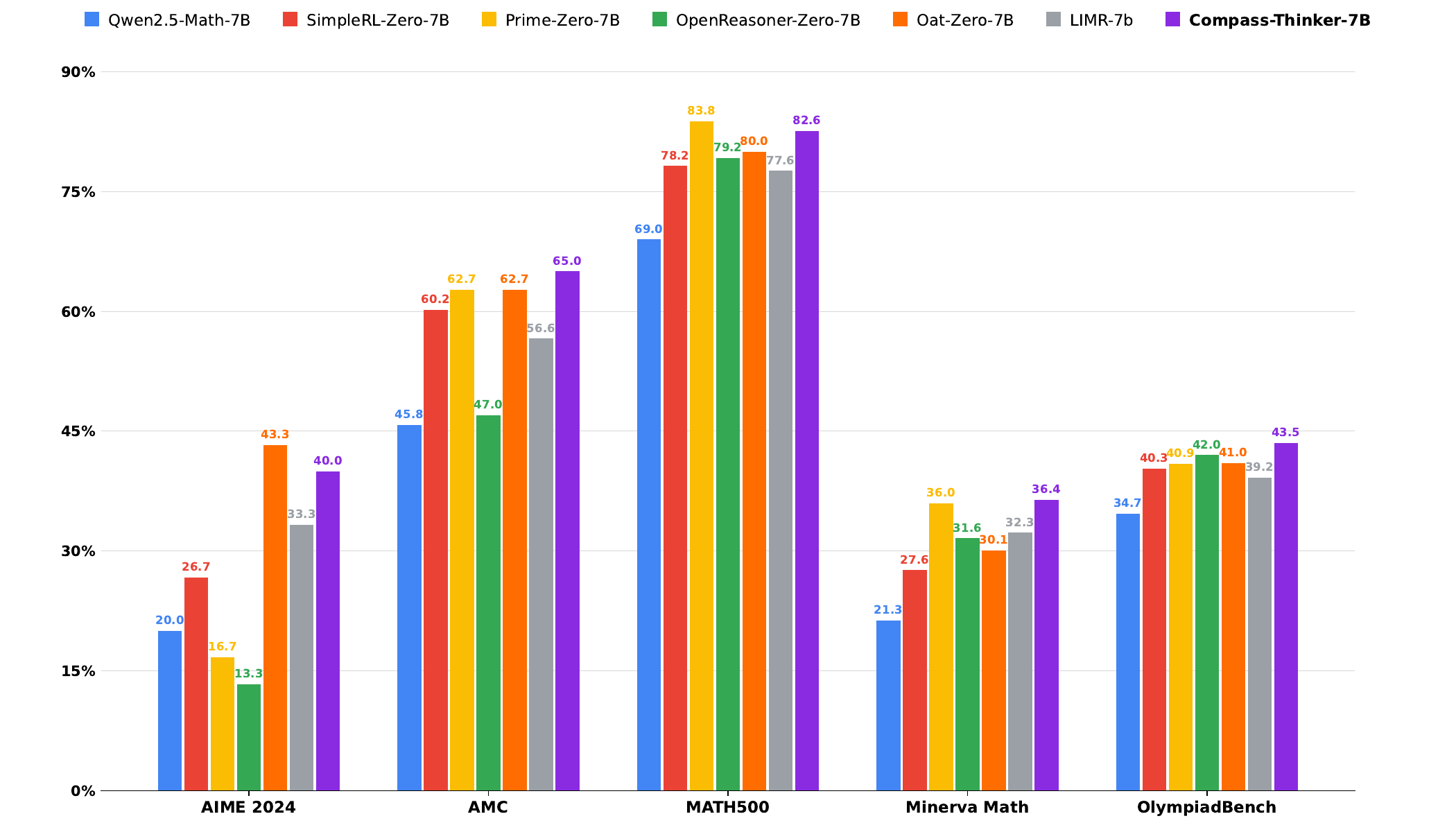}
    \caption{Comparison of Model Performance on Math Benchmarks. Compass-Thinker-7B achieved highly competitive performance in multiple mathematical benchmarks}
    \label{fig:compass-thinker-7b-main-result}
\end{figure}

\section{Introduction}

Large Language Models (LLMs) have seen significant advancements in recent years~\citep{brown2020languagemodelsfewshotlearners,bubeck2023sparksartificialgeneralintelligence,chowdhery2022palmscalinglanguagemodeling,touvron2023llamaopenefficientfoundation}, becoming foundational technologies across a wide range of natural language process (NLP) applications such as dialogue systems, code generation, and document understanding. Breakthrough models including GPT series~\citep{Achiam2023GPT4TR}, Gemini 2.5~\citep{comanici2025gemini25pushingfrontier}, LLaMA series~\citep{Dubey2024TheL3}, Qwen series~\citep{Yang2024Qwen25TR}, DeepSeek~\citep{deepseekai2025deepseekr1incentivizingreasoningcapability} series have pushed the boundaries of LLM by scaling up model size, training data and more advanced training paradigms.

Reinforce Learning from Human Feedback (RLHF) is an indispensable key point in LLM Post-Training process~\citep{stiennon2022learningsummarizehumanfeedback,christiano2023deepreinforcementlearninghuman,bai2022constitutionalaiharmlessnessai}. It is primarily used to align LLMs with human preferences and values, ensuring that LLMs are helpful, honest, and harmless~\citep{huang2025positionneedadaptiveinterpretation}. The classic RLHF method is a multi-stage process that typically involves learning a reward model from preference data~\citep{stiennon2022learningsummarizehumanfeedback,christiano2023deepreinforcementlearninghuman}, then optimizing the policy model to maximize that reward. Due to its complex modeling and training process, RLHF still faces many challenges in practical applications, and researchers have been studying and exploring simpler, more stable, and more efficient algorithms. Direct Preference Optimization (DPO)~\citep{rafailov2024directpreferenceoptimizationlanguage} algorithm reparameterizes the reward function in RLHF, enabling it to directly learn the policy model from preference data without requiring an explicit reward model. Group Relative Policy Optimization (GRPO)~\citep{shao2024deepseekmathpushinglimitsmathematical} algorithm improves training efficiency by optimizing memory usage in RLHF.

With the open-sourcing of DeepSeek-R1 models~\citep{deepseekai2025deepseekr1incentivizingreasoningcapability}, it has been shown that rule-based rewards can guide large models to achieve significant performance gains in complex tasks such as mathematical reasoning and code generation. 
Recent studies about R1-Zero-like continue to unfold, but most successful work relies on relatively large base models, e.g., 32B models. It is generally believed that it is challenging to achieve improvements in areas such as mathematics and code through RL in small models. Nevertheless, we believe that the effectiveness of RL training reasoning models depends on the inherent reasoning potential of the base model, and that significant improvements can still be achieved in relatively smaller models through proper planning and training.

Therefore, we introduce the Compass-Thinker-7B model. Compass-Thinker-7B is trained from an open-source base model through a specially designed RL Pipeline.  
We performed meticulous preprocessing on various open-source datasets, including data deduplication, noise filtering, low-quality filtering, and difficulty grading, ultimately constructing a dataset of 30k verifiable mathematics problems for the Reinforcement Learning Pipeline.
Evaluations show that Compass-Thinker-7B possesses exceptional reasoning potential and achieves superior performance on mathematics compared to the same-sized RL model.

\section{Preliminary}

\subsection{Proximal Policy Optimization}
Proximal Policy Optimization (PPO) ~\citep{schulman2017proximalpolicyoptimizationalgorithms}is a classic policy gradient reinforcement learning algorithm that stabilizes the training process by limiting the magnitude of policy updates. It uses a clipping mechanism when updating the policy to ensure that the change between the new and old policies does not exceed a predetermined range. This method improves the stability and efficiency of policy optimization while maintaining the simplicity, practicality, and sample efficiency of the algorithm. Specifically, PPO updates the policy by maximizing the following objectives:

$$
\Im_{PPO} = \mathbb{E}_t \left[ \min \left( r_t(\theta) \hat{A}_t, \text{clip}(r_t(\theta), 1 - \epsilon, 1 + \epsilon) \hat{A}_t \right) \right]
$$

where \( r_t(\theta) = \frac{\pi_\theta(a_t | s_t)}{\pi_{\theta_{\text{old}}}(a_t | s_t)} \) is the probability ratio of the policy at time step \( t \) .
\(\hat{A}_t\) is the Generalized Advantage Estimation (GAE)~\citep{schulman2018highdimensionalcontinuouscontrolusing} .
\(\epsilon\) is a small hyperparameter that limits the extent of policy updates.
\(\text{clip}(\cdot)\) function restricts \( r_t(\theta) \) to the range \([1 - \epsilon, 1 + \epsilon]\).This approach allows PPO to effectively control the magnitude of policy changes during optimization, maintaining stability in learning.

Compared to ordinary supervised learning, the training process of PPO is more complex and consumes more computational resources. The PPO training process typically involves loading four models:
\begin{itemize}
    \item \textbf{Policy Model}: The model to be optimized, which is also used to generate training samples.
    
    \item \textbf{Critic Model}: Estimates the current state value of the policy, used to calculate the advantage function and guide policy updates.

    \item \textbf{Reward Model}: A model that provides reward signals during the training process, particularly important in tasks where the environment reward is not explicitly defined, such as text generation.

    \item \textbf{Reference Model}: A fixed model used to calculate the Kullback-Leibler divergence to measure the deviation between the current policy model and the initialised policy, ensuring that the training results do not diverge excessively. 
\end{itemize}

Recent research has focused on reducing the excessive resource consumption in the PPO training process. For example, GRPO~\citep{shao2024deepseekmathpushinglimitsmathematical}, ReMax~\citep{li2024remaxsimpleeffectiveefficient}, and REINFORCE++~\citep{hu2025reinforceefficientrlhfalgorithm} eliminate the Critic Model and optimise training stability.

\subsection{Group Relative Policy Optimization}
Group Relative Policy Optimization (GRPO)~\citep{shao2024deepseekmathpushinglimitsmathematical} eliminates the Critic Model and estimates the advantage using a group-relative manner. 
Specifically, for a specific query $q$, a set of independent responses ${\{o_i,o_2,...,o_G\}}$ is sampled using the policy $\pi$, and the relative advantage of each response is calculated as follows:
$$
A_{i,j}=\frac{r_i-mean(\{r_i\}^{G}_{i=1})}{std(\{r_i\}^{G}_{i=1})}
$$

Similar to PPO, GRPO adopts a clipped objective, together with a directly imposed KL penalty term:
$$
\Im_{GRPO} =\mathbb{E}_{q\sim D,\{o_i\}^{G}_{i=1}\sim \pi_{\theta}}
[\frac{1}{G}\displaystyle \sum^G_{i=1}\frac{1}{|o_i|}\sum^{|o_i|}_{t=1}
min(\frac{\pi_{\theta}(o_i|q)}{\pi_{old}(o_i|q)}A_{i,j}, 
clip(\frac{\pi_{\theta}(o_i|q)}{\pi_{old}(o_i|q)}, 1-\epsilon_{low}, 1+\epsilon_{high})A_{i,j})]
$$

\subsection{Dynamic Sampling Policy Optimization}
Based on the GRPO algorithm, Dynamic sAmpling Policy Optimization (DAPO)~\citep{yu2025dapoopensourcellmreinforcement} algorithm introduces multiple innovations to improve the algorithm performance:
$$
\Im_{DAPO} =\mathbb{E}_{q\sim D,\{o_i\}^{G}_{i=1}\sim \pi_{\theta}}[\frac{1}{\sum^{G}_{i=1}|o_i|}\displaystyle \sum^{G}_{i=1} \sum^{|o_i|}_{j=1} 
min(\frac{\pi_{\theta}(o_i|q)}{\pi_{old}(o_i|q)}A_{i,j}, 
clip(\frac{\pi_{\theta}(o_i|q)}{\pi_{old}(o_i|q)}, 1-\epsilon_{low}, 1+\epsilon_{high})A_{i,j})]
$$

\begin{itemize}
    \item The Clip-Higher strategy alleviates the problem of entropy collapse during RL  training.
    \item Dynamic sampling improves sample efficiency.
    \item Token-Level Policy Gradient Loss alleviates the unhealthy increase in response length.
    \item Overlong Reward Shaping for truncated samples can reduce noise and significantly stabilizes training and enhances performance.
\end{itemize}

\subsection{Rule-Based Reward Modeling}
In classic RLHF, preference data is usually used to pre-train a reward model based on a large model. The reward model can give a reward value for a query and response. However, this type of reward model may encounter the problem of reward hacking during use~\citep{shihab2025detectingmitigatingrewardhacking,gao2022scalinglawsrewardmodel}. Instead, we can directly use the final accuracy of a verifiable task as the outcome reward, computed using the following rule:
\begin{equation}
\label{eq6}
R(\hat{y},y)=\left\{
\begin{aligned}
1 & , & is\_equivalent(\hat{y},y) \\
-1 & , & otherwise
\end{aligned}
\right.
\end{equation}
where ${y}$ is the ground-truth answer and ${\hat{y}}$ is the predicted answer. This is proved to be an effective approach to activating the base model’s reasoning capability, as shown in multiple domains such as automated theorem
proving, computer programming, and mathematics competition.

\section{Compass-Thinker-7B}

For the RL algorithm, we applied a variant of the GRPO~\citep{shao2024deepseekmathpushinglimitsmathematical} algorithm, combining many improvements from the industry. 
For each query ${q}$, the algorithm uses the old policy $\pi_{old}$ to sample a group of responses${\{o_i,o_2,...,o_G\}}$. Each response will get a corresponding reward $\{r_1, r_2,...,r_G\}$ from the Reward Function.
The policy $\pi_{\theta}$ will be updated by maximizing the following objective:

$$
\Im_{GRPO} =\mathbb{E}_{q\sim D,\{o_i\}^{G}_{i=1}\sim \pi_{\theta}}[\frac{1}{\sum^{G}_{i=1}|o_i|}\displaystyle \sum^{G}_{i=1} \sum^{|o_i|}_{j=1} 
min(\frac{\pi_{\theta}(o_i|q)}{\pi_{old}(o_i|q)}A_{i,j}, 
clip(\frac{\pi_{\theta}(o_i|q)}{\pi_{old}(o_i|q)}, 1-\epsilon_{low}, 1+\epsilon_{high})A_{i,j})]
$$

where $\epsilon_{low}$ and $\epsilon_{high}$ are hyper-parameters. $A_{i,j}$ is the advantage, which is computed by the rewards
$\{r_1, r_2,...,r_G\}$ of responses in the same group:
$$
A_{i,j}=\frac{r_i-mean(\{r_i\}^{G}_{i=1})}{std(\{r_i\}^{G}_{i=1})}
$$

Upon on the original GRPO algorithm, we also introduced some important improvements from recent research:

\begin{itemize}
    \item \textbf{Remove KL Loss}: Directly removing the kl loss in the GRPO objective function can effectively unleash the potential of the model, and numerous experiments have shown that removing the kl loss does not compromising training stability.
    
    \item \textbf{Dynamic Sampling}: In the sampling phase, we filter out samples with accuracy rates of 0 or 1 within the group, as these samples do not produce effective gradients because their advantage is 0.

    \item \textbf{Clip-Higher}: We increase the upper clip bounds $\epsilon_{high}$, with a fixed lower clip bounds $\epsilon_{low}$. It can effectively alleviate the problem of model entropy collapse during RL training and ensure the diversity of sampling results.
    
\end{itemize}

\section{Experiment}

\subsection{Implementation}
We implement Reinforcement Learning Pipeline using the verl framework~\citep{Sheng_2025}. verl is a flexible, efficient and production-ready RL training library for large language models (LLMs). verl decouples computation and data dependencies, enabling seamless integration with existing LLM frameworks, such as FSDP~\citep{zhao2023pytorchfsdpexperiencesscaling}, Megatron-LM~\citep{shoeybi2020megatronlmtrainingmultibillionparameter}, vLLM~\citep{kwon2023efficientmemorymanagementlarge}, SGLang~\citep{zheng2024sglangefficientexecutionstructured}. it supports various placement of models onto different sets of GPUs for efficient resource utilization and scalability across different cluster sizes.

\textbf{Hyperparameters} 
In our experiment, we employed a training batch size of 128, with an actor mini-batch size
of 32. We executed 16 gradient updates per training iteration at a learning rate of 1e-6. The
maximum sequence length was set to 4096 tokens to facilitate complex reasoning tasks. During
the training phase, both temperature and top-p parameters were configured at 1.0 to promote
output diversity.

\textbf{Base Models} We took Qwen2.5-7B-Math~\citep{qwen2025qwen25technicalreport} as the starting point of Compass-Thinker 7B for RL pipeline training. To study the performance of the base model at different stages after RL, we also adopted different base models and executed the same RL pipeline.

\subsection{Dataset}
We collected a training set of prompts for RL training from multiple open-source datasets.
such as NuminaMATH~\citep{li2019numinamath}.When collecting mathematical data, we first ensured that every prompt in the data had a verifiable ground truth.Secondly, we filtered out noisy data, which included web links or image links inaccessible to the model, potentially leading to unexpected output results.
Additionally, the collected mathematical data included a portion of ungraded data. To meet the requirements of course reinforcement learning, we used a method similar to Big-Math to grade this data based on a specific LLM. We used Llama-8b to label this dataset and categorized it into difficulty levels 1–5 based on accuracy percentiles.

In our testing, we found that the Math-Verify library had performance issues and some bugs, so we developed a separate verification program tailored for the mathematical domain, which better meets our requirements for verification speed. o ensure verification accuracy, we used the verification program to filter the training data, removing data that could not be identified or failed verification.

\subsection{Evaluation}
In order to comprehensively evaluate model performance, we evaluate our models on a series of challenging benchmark datasets:

\begin{itemize}
    \item \textbf{American Invitational Mathematics Examination 2024(AIME2024)}~\citep{aime2024}: A challenging mathematical reasoning competition dataset containing 30 questions with integer answers,designed to assess precise mathematical reasoning abilities.

    \item \textbf{American Mathematics Competitions(AMC)}~\citep{li2019numinamath}: One of the most influential high school mathematics competitions in the United States and globally,the competition covers fundamental mathematical knowledge, including algebra, geometry, number theory, and combinatorial mathematics, with 83 challenging questions.
    
    \item \textbf{Math 500}~\citep{lightman2023letsverifystepstep}: A benchmark dataset comprising 500 math problems,used to evaluate the performance of large language models on mathematical reasoning tasks. Specifically, the MATH 500 dataset includes math problems ranging from middle school to high school difficulty levels, covering multiple fields such as algebra, geometry, probability, and statistics.

    \item \textbf{Minerva Math}~\citep{lewkowycz2022solvingquantitativereasoningproblems}: This dataset includes 272 questions, covering a wide range of difficulty levels in mathematics, from basic arithmetic to complex algebraic and geometric problems, aiming to comprehensively evaluate the model's performance in the field of mathematics.
    
    \item \textbf{Olympiad Bench}~\citep{he2024olympiadbenchchallengingbenchmarkpromoting}: It is a high-difficulty evaluation set focused on assessing the reasoning capabilities of large language models and large multimodal models in complex mathematical, physical, and other scientific fields. The questions are selected from international mathematics/physics olympiads (IMO/IPhO), the Chinese college entrance examination, and globally renowned qualifying competitions.
\end{itemize}

\subsection{Results}
We conducted large-scale reinforcement learning training based on the Qwen2.5-Math-7B model and performed comprehensive evaluations in the field of mathematics. 
Table\ref{Table:Main Results} shows the evaluation results in various mathematical benchmark tests.  
Compared to the Base Model, significant improvements were achieved in all benchmark evaluations after training through our reinforcement learning recipe.
Especially in the challenging AIME24 evaluation, accuracy improved from $20\%$ to $40\%$, demonstrating that our RL recipe can more fully unlock the model's potential.
Compared to other RL models of same size, Compass-Thinker-7B achieved the highest scores in AIME24, AMC, Minerva Math, and Olympiad Bench.
This indicates the effectiveness of our data construction and algorithm optimization in the RL phase.

\begin{table}[H]
\caption{ Evaluation results in multiple mathematical benchmark.Comparison with RL models of the same size,Compass-7B-Thinker has achieved highly competitive performance.}
\resizebox{\columnwidth}{!}{
\begin{tabular}{@{}lccccc@{}l}

\toprule
\textbf{}            & \textbf{AIME2024} & \textbf{AMC} & \textbf{Math500} &  \textbf{Minerva Math} &\textbf{Olympiad Bench}  \\ \midrule
Qwen2.5-Math-7B&         20.0&    45.8&   69.0&   21.3&  34.7  \\ 
SimpleRL-Zero-7B&        26.7&    60.2&   78.2&   27.6&  40.3  \\ 
Prime-Zero-7B&           16.7&    62.7&   \textbf{83.8}&   36.0&  40.9  \\ 
OpenReasoner-Zero-7B&    13.3&    47.0&   79.2&   31.6&  42.0  \\ 
Oat-Zero-7B&             \textbf{43.3}&    62.7&   80.0&   30.1&  41.0  \\ 
LIMR-7b&                 33.3&     56.6&   77.6&   32.3&  39.2  \\ 
\toprule
\textbf{Compass-Thinker-7B}& 40.0& \textbf{65.0}&   82.9&   \textbf{36.4}&  \textbf{43.5}\\ 
\bottomrule
\end{tabular}
}
\label{Table:Main Results}
\end{table}

\section{Conclusion}
In this paper, we propose the Compass-Thinker-7B model, which aims to explore the potential of reinforcement learning with less computational resources and costs.After training with our RL recipe, Compass-Thinker-7B achieved highly competitive performance in multiple mathematical benchmarks, with particularly notable improvements in the challenging AIME24 evaluation.This demonstrates that our meticulously designed RL pipeline can fully unlock the model's potential.We hope this work can offer insights for developing more powerful reasoning models.
\bibliography{main}

\section*{Contributors}

Anxiang Zeng \\
Haibo Zhang \\
Kaixiang Mo \\
Long Zhang \\
Shuman Liu \\
Yanhui Huang \\
Yawen Liu \\
Yuepeng Sheng \\
Yuwei Huang

Within each role, authors are listed alphabetically by the first name.

\end{document}